\documentclass[11pt]{article}

\usepackage[final]{acl}

\usepackage{times}
\usepackage{latexsym}

\usepackage[T1]{fontenc}

\usepackage[utf8]{inputenc}

\usepackage{microtype}

\usepackage{inconsolata}

\usepackage{url}
\usepackage{amsmath}
\usepackage{algorithm}
\usepackage{algpseudocode}
\usepackage{graphicx}
\usepackage[table]{colortbl}
\usepackage{tabularx}
\usepackage[most]{tcolorbox}
\usepackage{hyperref}
\usepackage{booktabs} 
\usepackage{multirow} 

\usepackage{setspace} 
\algrenewcommand\alglinenumber[1]{\scriptsize #1} 
\algrenewcommand\algorithmicindent{1.2em} 

\makeatletter
\renewcommand{\ALG@beginalgorithmic}{\setstretch{0.95}} 
\makeatother

\title{MaiBERT: A Pre-training Corpus and Language Model for Low-Resourced Maithili Language}

\author{Sumit Yadav\textsuperscript{1}, Raju Kumar Yadav\textsuperscript{1}, Utsav Maskey\textsuperscript{2}, Gautam Siddharth Kashyap\textsuperscript{2}, \AND
        Ganesh Gautam\textsuperscript{1}, Usman Naseem\textsuperscript{2}    \\
        \textsuperscript{1}IOE, Pulchowk Campus, Lalitpur, Nepal \\
        \texttt{\{076bct088.sumit, 076bct100.raju, ganesh\}@pcampus.edu.np} \\
        \textsuperscript{2}Macquarie University, Sydney, Australia \\
        \texttt{\{utsav.maskey,\allowbreak usman.naseem\}@mq.edu.au} \\
        \texttt{gautam.kashyap@hdr.mq.edu.au} \\
        }

\begin{document}
\maketitle
\begin{abstract}
Natural Language Understanding (NLU) for low-resource languages remains a major challenge in NLP due to the scarcity of high-quality data and language-specific models. Maithili, despite being spoken by millions, lacks adequate computational resources, limiting its inclusion in digital and AI-driven applications. To address this gap, we introduce \textbf{maiBERT}, a BERT-based language model pre-trained specifically for Maithili using the Masked Language Modeling (MLM) technique. Our model is trained on a newly constructed Maithili corpus and evaluated through a news classification task. In our experiments, \textbf{maiBERT} achieved an accuracy of 87.02\%, outperforming existing regional models like NepBERTa and HindiBERT, with a 0.13\% overall accuracy gain and 5–7\% improvement across various classes. We have open-sourced \textbf{maiBERT} on Hugging Face\footnote{\url{https://huggingface.co/rockerritesh/maiBERT_TF}},
enabling further fine-tuning for downstream tasks such as sentiment analysis and Named Entity Recognition (NER).
\end{abstract}

\section{Introduction}

\begin{figure}[t!]
    \centering
    \includegraphics[width=\columnwidth]{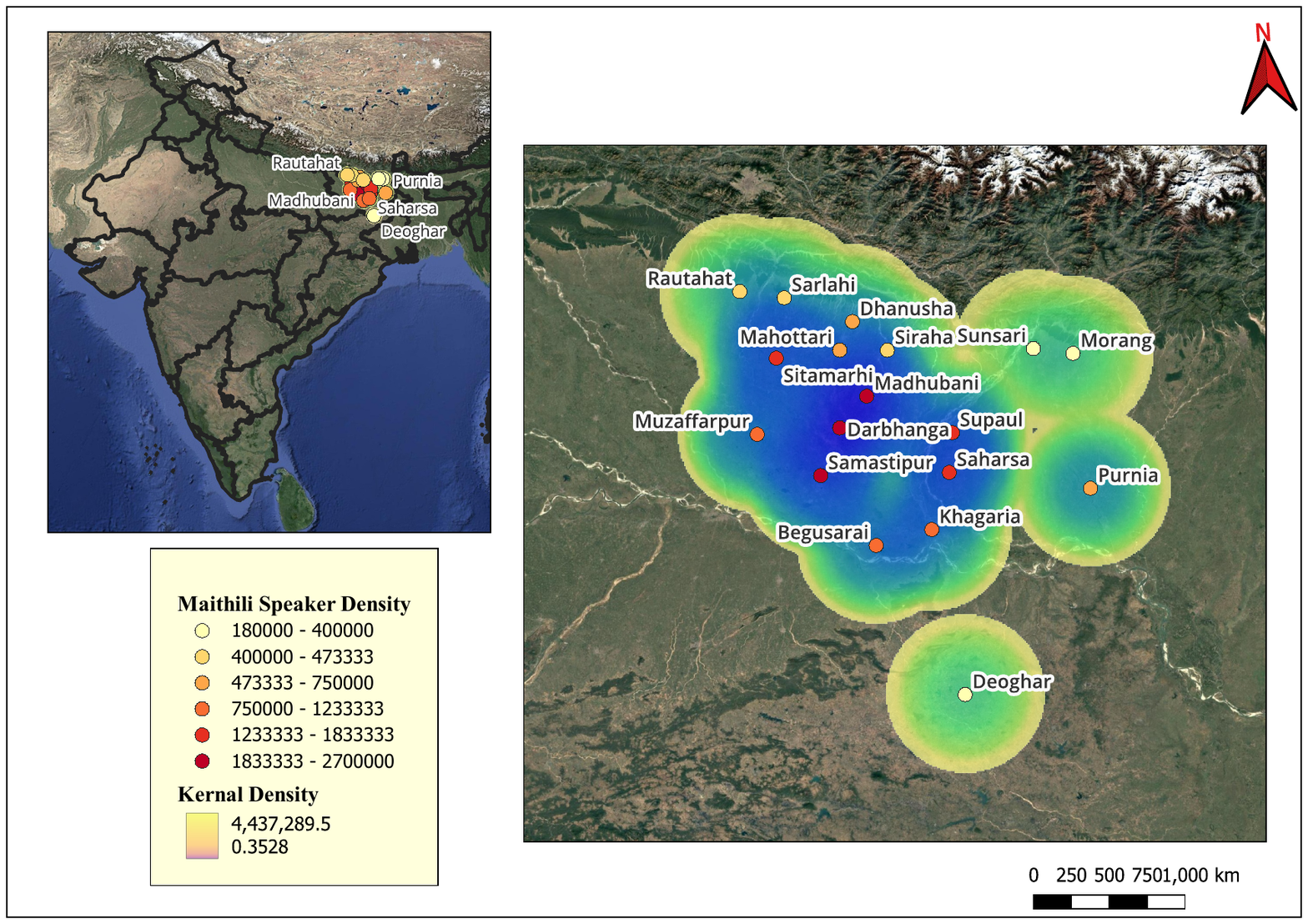}
    \caption{Density-Based Map of Maithili Language.}
    \label{fig1}
\end{figure}

Natural Language Processing (NLP) has seen rapid advancements in recent years, with the development of sophisticated models like BERT \citep{devlin2019bert}, GPT \citep{radford2018improving}, and T5 \citep{raffel2020exploring} that enable machines to understand, interpret, and generate human language. However, most of this progress has primarily focused on high-resource languages such as English, Mandarin, and Hindi. Low-resource languages—languages that lack sufficient digital content, annotated corpora, and computational resources—have remained largely underserved. One such language is Maithili\footnote{In terms of linguistic structure, Maithili follows a syntax and morphology that are comparable to other Devanagari-script languages such as Hindi and Nepali. It comprises 47 segmental phonemes: 8 vowel phonemes characterized by duration and quality, 39 consonant phonemes, and 10 numerals represented as letters.} as shown in Fig. \ref{fig1}, an Indo-Aryan language spoken by millions of people in Nepal and India, particularly in the Terai region of Nepal (mainly Madhesh Province and parts of Koshi Province) and neighboring districts in the Indian state of Bihar.

Maithili is constitutionally recognized in Nepal and India and is widely spoken in both formal and informal contexts. Despite its rich linguistic and cultural heritage, Maithili has received minimal attention in the computational linguistics community \citep{mundotiya2021linguistic}. Unlike related languages such as Hindi and Nepali, which have been the subject of substantial research \citep{joshi2022l3cube, pudasaini2023nepalibert, maskey-etal-2022-nepali} and have several publicly available resources \citep{bal2024strategies, bafna2022subword}, Maithili lacks a comprehensive digital footprint. There are no large-scale pre-trained language models, standardized datasets, or open-source tools dedicated to this language, which significantly impedes the development of NLP applications for Maithili speakers. While existing models developed for related languages such as \citep{hartmann2017portuguese, alian2020semantic, pandit2019improving, patil2022supervised, nie2022cross, ramchandra2019deep, lalrempuii2021improved, priyadarshi2020towards, mundotiya2022hierarchical} have shown some promise, they fall short when applied directly to Maithili due to dialectal variations, unique lexical items, and morphological structures that distinguish it from other Indo-Aryan languages.

To bridge this critical gap, we propose a Maithili-specific language model using the BERT architecture. Our work begins with the construction of a high-quality Maithili corpus collected from diverse sources, including online articles, books, and conversational data. This raw corpus forms the basis for pre-training a Masked Language Model (MLM) tailored specifically for Maithili. Following pre-training, we fine-tune the model for a downstream NLP task—text classification. Our model, referred to as \textbf{maiBERT}, is then benchmarked against State-of-the-Art (SOTA) to evaluate performance.

\section{Related Works}

NLP has shown remarkable advancements in foundational tasks such as semantic similarity and text categorization, significantly benefiting high-resource languages but posing persistent challenges for low-resource languages like Maithili \citep{chandrasekaran2021evolution, zhang2015survey}. 
Therefore, to address the data scarcity in low-resource languages, transfer learning has emerged as a widely adopted strategy, where pretrained models from high-resource languages are fine-tuned on target languages \citep{ali2022hate}. \citet{patil2022supervised} demonstrated this by adapting an English sentiment model for Marathi through parameter optimization, although their study lacks scalability for languages with significantly different syntax and morphology. Similarly, \citet{nie2022cross} proposed a novel cross-lingual emotion categorization model for Swahili and Urdu, but the absence of domain-specific datasets restricts real-world applicability. \citet{ramchandra2019deep} explored semantic similarity in Hindi using multilingual deep learning models such as BERT and LASER, highlighting their comparative effectiveness; however, their results do not consider dialectal or regional variations commonly found in Hindi’s linguistic relatives. \citet{lalrempuii2021improved} developed a NER framework for Assamese using manually annotated data and neural networks, achieving reasonable accuracy; however, the manual annotation process is labor-intensive and limits scalability. \citet{priyadarshi2020towards} created a Maithili POS tagger using CRF and neural word embeddings, reaching 85.88\% accuracy with a 52,190-word corpus, but the system’s performance may degrade for out-of-vocabulary or code-switched text. \citet{mundotiya2022hierarchical} developed annotated corpora and POS/chunking tools for Bhojpuri, Maithili, and Magahi, showing high accuracy for SAHBiLC and FineSAHBiLC models (e.g., 0.86\%–0.95\% for POS and chunking); still, their approach is limited by the static nature of the corpus and its lack of domain diversity. Word vectorization techniques, in general, have proven to be transformational for low-resource languages, yet the challenge remains in crafting embeddings that reflect the linguistic and morphological intricacies of underrepresented languages without relying excessively on sister-language transfer models. While transfer learning and distributional models enable rapid development, they often inherit biases and linguistic mismatches from their source languages.


\section{Dataset}

\subsection{Pre-training Corpus:} 

With the objective of training language models for Maithili,  we gathered approximately 18 million words 
  by combining and de-duplicating data from three sources: (1) Maithili Wikipedia articles\footnote{\url{https://mai.wikipedia.org}} \textasciitilde880K
   words), (2) Digitized books\footnote{\url{https://archive.org/details/maithili-books/}} obtained through OCR processing of 418 scanned documents sourced from
  arXiv and other archives (\textasciitilde10.4 million words), and (3) raw newspaper articles, such as Esamaad\footnote{\url{https://esamaad.com/}}, Maithilijindabad\footnote{\url{https://maithilijindabaad.com/}}, and Maithili Purnajagran Prakash\footnote{\url{https://mppdainik.com/}} (\textasciitilde6.6 million words).
  The OCR extraction was performed using Tesseract \cite{4376991} on Maithili literary works and historical texts. After
   preprocessing and merging, the final corpus comprises over 450MB of cleaned Maithili text, making it
  one of the largest publicly compiled datasets for this language.

\begin{figure}[h!]
    \centering
    \includegraphics[width=\columnwidth]{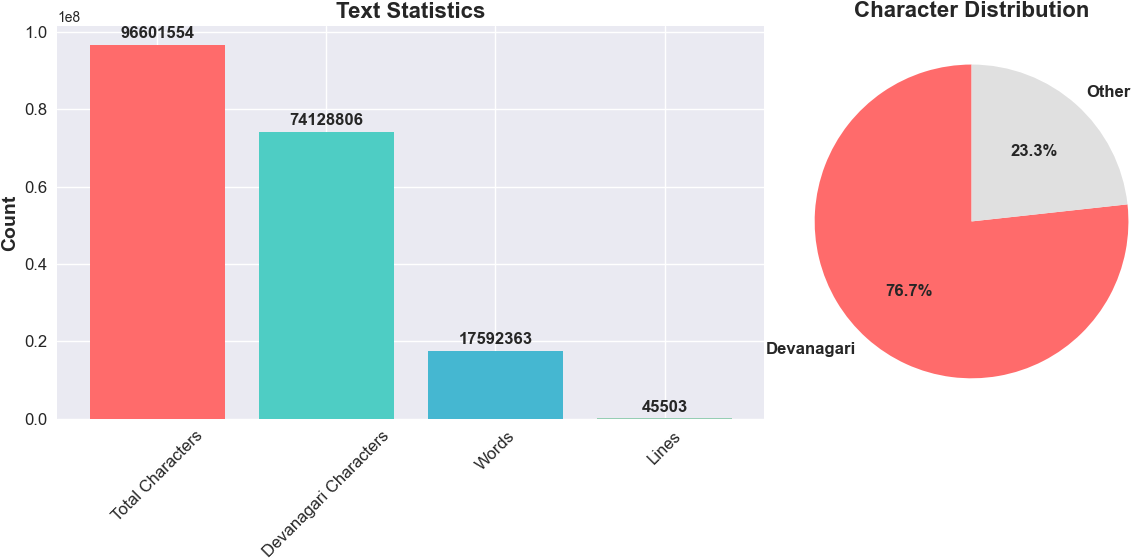}
    \caption{Comprehensive text statistics of the Maithili corpus showing character distribution, word counts, and linguistic features.}
    \label{fig:text_data_stats}
    \vspace{-0.45cm}
\end{figure}

The combined raw corpus\footnote{\url{https://huggingface.co/datasets/rockerritesh/maithiliNewsData}}
for MLM pre-training consists of 1,028,017 sentences, with an average sentence length of 16.88 words and an average word length of 4.44 characters.
Figure~\ref{fig:text_data_stats} provides an overview of the Maithili text corpus, illustrating how characters, words, and key linguistic features are distributed across the dataset.

\noindent The corpus has a rich vocabulary of 1,357,081 unique tokens, highlighting its lexical diversity. A linguistic analysis revealed frequent occurrences of the Maithili-specific punctuation marker, with a total of 1,044,037 appearances across all text files—providing insights into syntactic structure and sentence boundaries.

\begin{figure}[hpb!]
    \centering
    \includegraphics[width=0.9\columnwidth]{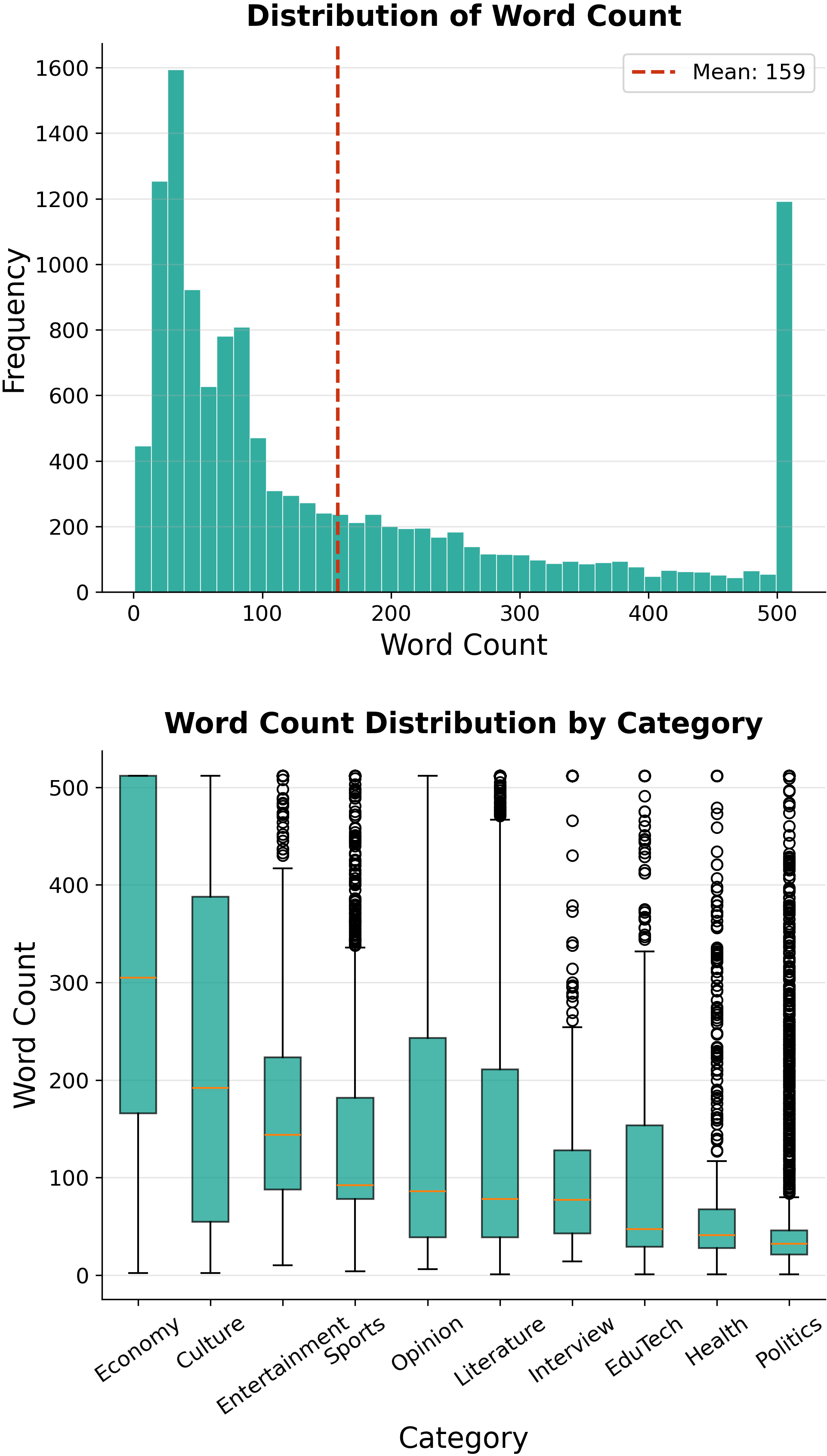}
    \caption{Word distribution patterns in Maithili corpus.}
    \label{fig:word_distribution}
\end{figure}

Figure~\ref{fig:word_distribution} displays the overall word distribution patterns in the Maithili text corpus.

\noindent For enhanced language modeling, the raw data underwent extensive preprocessing, including normalization, lemmatization, and tokenization using a customized BertTokenizer\footnote{\url{https://www.tensorflow.org/text/api_docs/python/text/BertTokenizer}}, which preserved linguistic features specific to Devanagari script while filtering out non-Devanagari characters.

\textbf{Classification Dataset:} For downstream evaluation, we constructed a separate labeled test set from the Maithili news sources, as mentioned in the pre-training corpus.

\begin{figure}[h!]
    \centering
    \includegraphics[width=0.9\columnwidth]{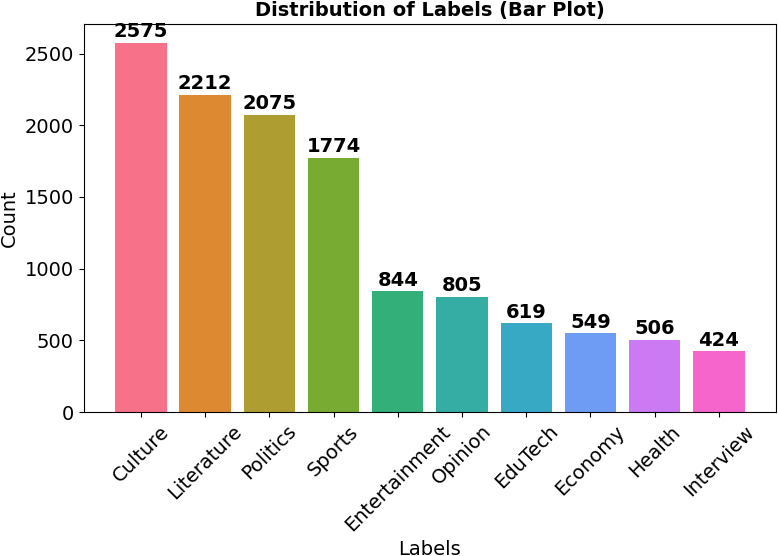}
    \caption{Distribution of news categories in the Maithili classification dataset.}
    \label{fig:category_distribution}
\end{figure}

The annotation process involved 3 native speakers and 2 linguistic experts who labeled the corpus into ten predefined categories, totaling 11,959 labeled instances, as shown in Figure~\ref{fig:category_distribution}. This distribution reflects the natural occurrence of topics in Maithili news media, with cultural and literary content being predominant.

\begin{figure}[h!]
    \centering
    \includegraphics[width=0.9\columnwidth]{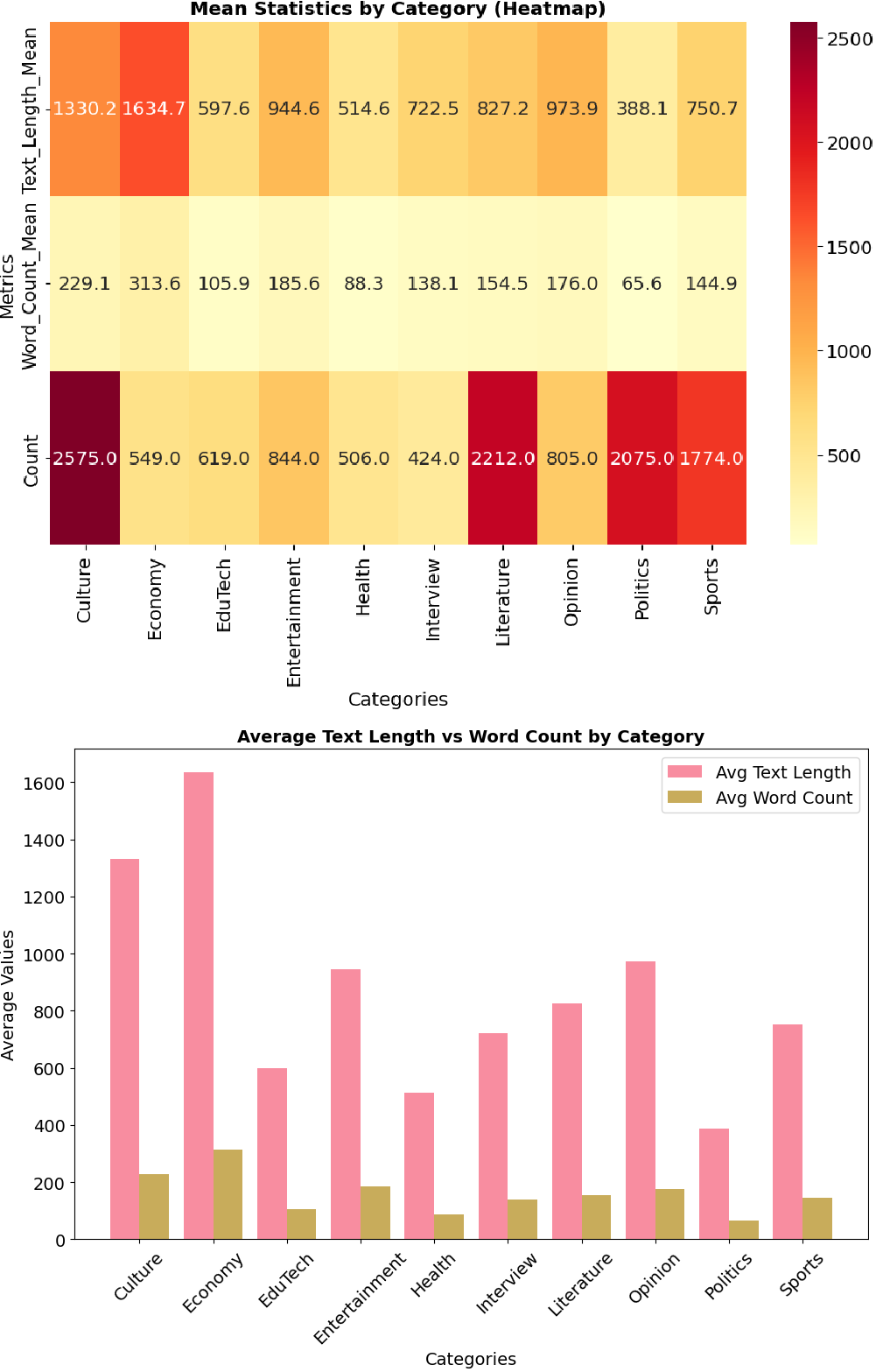}
    \caption{Word count distribution across different news categories in the classification dataset.}
    \label{fig:words_by_category}
\end{figure}

\noindent To illustrate the varying levels of content complexity across categories, Figure~\ref{fig:words_by_category} presents the distribution of word counts for different news sections.

\begin{figure}[h!]
    \centering
    \includegraphics[width=0.9\columnwidth]{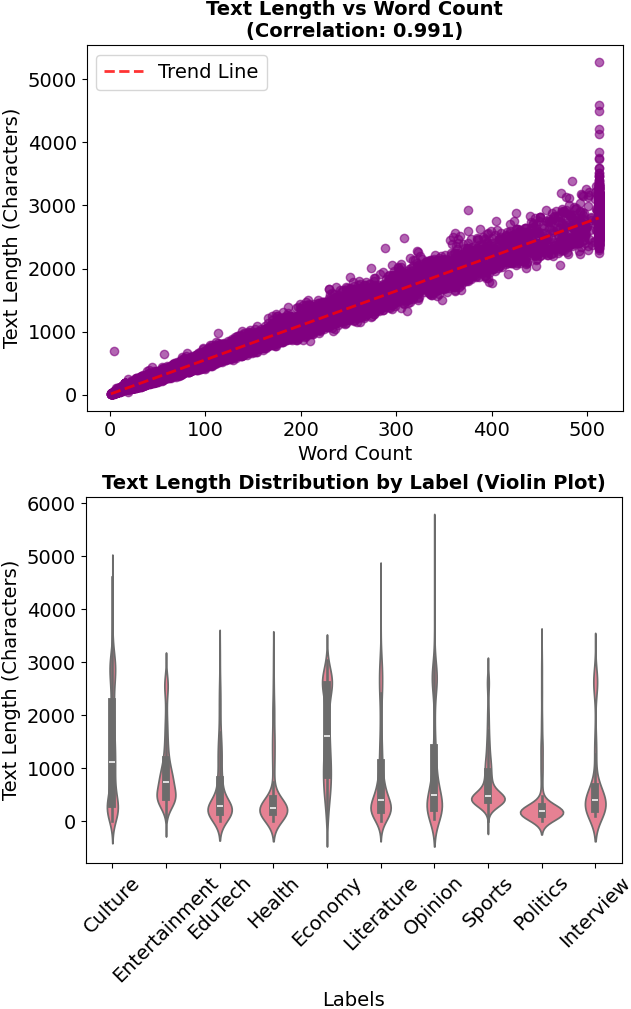}
    \caption{Distribution of text lengths in the Maithili classification dataset.}
    \label{fig:text_length}
\end{figure}

\noindent Figure~\ref{fig:text_length} shows the distribution of text lengths across the classification dataset, illustrating the variety in article lengths.

\begin{table}[t]
\centering
\small
\begin{tabular}{l|rrr}
\toprule
\textbf{Category} & \textbf{Count} & \textbf{Avg Words} & \textbf{Avg Sent} \\
\midrule
Culture & 2,575 & 147.4 & 8.5 \\
Economy & 549 & 176.8 & 8.5 \\
EduTech & 619 & 57.0 & 6.7 \\
Entertainment & 844 & 156.1 & 8.1 \\
Health & 506 & 58.5 & 4.3 \\
Interview & 424 & 95.4 & 6.8 \\
Literature & 2,212 & 155.0 & 10.3 \\
Opinion & 805 & 123.1 & 9.9 \\
Politics & 2,075 & 52.1 & 4.8 \\
Sports & 1,774 & 135.7 & 8.1 \\
\midrule
\textbf{Total} & \textbf{12,383} & \textbf{130.7} & \textbf{7.6} \\
\bottomrule
\end{tabular}
\caption{Dataset overview: Distribution and statistics across text categories for Count, Average number of words and Sentences.}
\label{tab:dataset_overview}
\end{table}

\noindent To validate the quality and consistency of annotations, inter-annotator agreement was computed using Cohen's Kappa coefficient, yielding a strong agreement score of 0.82, indicating reliable annotation practices. The labeled dataset was then split into training (70\%), validation (15\%), and test (15\%) subsets to ensure robust performance evaluation. Table~\ref{tab:dataset_overview} provides a comprehensive overview of the classification dataset characteristics across different text complexity levels.

\label{Dataset Analysis}

\section{Methodology}

In the development of \textbf{maiBERT}, a dedicated Transformer-based language model for the Mithila language, we adopt the TensorFlow implementation of the BERT\footnote{\url{https://huggingface.co/docs/transformers/en/model_doc/bert}} architecture, pretrained using the MLM objective. Given an input sequence \( \mathbf{x} = [x_1, x_2, \ldots, x_n] \), the model first maps each token \( x_i \) into a corresponding embedding \({e}_i \in {R}^d \), where \( d \) denotes the embedding dimension, forming an input matrix \({E} \in {R}^{n \times d} \). This matrix is combined with positional encodings \({P} \in {R}^{n \times d} \), ensuring sequential order preservation: \({Z}^{(0)} = {E} + {P} \). Each Transformer layer applies a multi-head self-attention mechanism, wherein attention heads \( h \) compute the scaled dot-product attention function as shown in Equation (1).
\begin{equation}
\text{Attention}({Q}, {K}, {V}) = \text{softmax}\left( \frac{{Q}{K}^\top}{\sqrt{d_k}} \right) {V}
\end{equation}
where query, key, and value matrices \({Q}, {K}, {V} \in {R}^{n \times d_k} \) are derived via learned linear projections. Multiple heads capture diverse syntactic and semantic patterns, concatenated and projected to form the output \({Z}^{(l+1)} \). The stacked encoder layers include residual connections and layer normalization, enabling gradient flow across layers and mitigating vanishing gradients. The MLM training objective stochastically masks a subset of tokens \( x_i \) in the input sequence and optimizes the Negative Log-Likelihood (NLL) of correctly predicting the masked tokens as shown in Equation (2).
\begin{equation}
\mathcal{L}_{\text{MLM}} = - \sum_{i \in \mathcal{M}} \log P(x_i \mid {x}_{\mathcal{M}})
\end{equation}
where \( \mathcal{M} \subseteq \{1, \ldots, n\} \) denotes the masked positions and \( \mathbf{x}_{\mathcal{M}} \) represents the input sequence with masked tokens. We train a Mithila language Language model with text corpus sourced from \texttt{MithilaTextCorpus-v1} (refer to Section~\ref{Dataset Analysis}), ensuring a representative distribution of regional grammar, morphology, and vocabulary richness. The corpus was tokenized using \texttt{SentencePiece-BPE}, a subword-level tokenizer based on Byte-Pair Encoding (BPE), yielding a vocabulary \( \mathcal{V} \) of size \texttt{30,000}, ensuring subword-level coverage of rare and compound words. Each training sample was tokenized into sequences of length \( n \leq 512 \), padded or truncated appropriately. The final vocabulary embedding matrix \({W}_e \in {R}^{|\mathcal{V}| \times d} = {R}^{30000 \times 768} \) was learned jointly with model parameters. Training was conducted over \texttt{1,000,000} steps with a batch size of \texttt{64} and a learning rate \( \eta = 5 \times 10^{-5} \), using the Adam optimizer with bias correction and weight decay regularization. Gradient updates followed the update rule according to Equation~(3):
\begin{equation}
\theta_{t+1} \leftarrow \theta_t - \eta \cdot \frac{\hat{m}_t}{\sqrt{\hat{v}_t} + \epsilon}
\end{equation}
where \( \hat{m}_t \) and \( \hat{v}_t \) are bias-corrected estimates of the first and second moments of gradients. The model was implemented using TensorFlow’s\footnote{\texttt{TFAutoModelForMaskedLM}} API, allowing compatibility with Hugging Face’s pretrained model interface. During inference or downstream fine-tuning, the model can be easily reloaded using TensorFlow’s transformer ecosystem. For text generation or completion, masked token positions are inferred using Maximum a Posteriori (MAP) estimates. For classification tasks, the output hidden state \({H} \in {R}^{n \times d} \) from the final encoder layer is pooled using the CLS token representation \({h}_{\text{[CLS]}} \), then passed through a task-specific classification head \({y} = \text{softmax}({W}_c{h} + {b}_c) \) with cross-entropy loss as shown in Equation (4).
\begin{equation}
\mathcal{L}_{\text{CE}} = -\sum_{i=1}^{C} y_i \log(\hat{y}_i)
\end{equation}
where \( C \) is the number of classes. Fine-tuning for sequence classification tasks employs the \texttt{TFAutoModelForSequenceClassification} interface, supporting task adaptation on custom Mithila language datasets. The model was trained on NVIDIA RTX 4090 GPUs with 24\,GB VRAM using mixed-precision training (FP16) to reduce memory usage and accelerate matrix operations. Training leveraged TensorFlow’s XLA compiler and \texttt{tf.keras.mixed\_precision} policy for automatic loss scaling and efficient GPU utilization, resulting in improved throughput and reduced training time without compromising model accuracy. To mitigate overfitting, dropout regularization \( \mathcal{D}(p = 0.1) \), early stopping with a patience of 3 epochs based on validation loss, and weight decay penalties \( \lambda \|\theta\|^2 \) with \( \lambda = 0.01 \) were applied. Token masking followed the BERT convention, with 15\% of tokens masked: 80\% replaced with \texttt{[MASK]}, 10\% replaced with a random token, and 10\% left unchanged. \textbf{maiBERT} embedding space offers potential for visualization and semantic similarity tasks, whereby cosine similarity \( \cos(\theta) = \frac{{a} \cdot {b}}{\|{a}\|\|{b}\|} \) is used to quantify proximity in the learned semantic space. Algorithm \ref{Alg} represents the entire flow of \textbf{maiBERT}.

\begin{algorithm*}[t!]
\setstretch{0.95} 
\caption{: \textbf{maiBERT}: Pretraining and Fine-tuning Procedure}
\small
\label{Alg}
\begin{algorithmic}[1]
\Require Corpus $\mathcal{C}$, vocabulary size $|\mathcal{V}| = 30000$, sequence length $n \leq 512$, embedding dimension $d = 768$, learning rate $\eta = 5 \times 10^{-5}$, batch size = 64, total steps $T = 1{,}000{,}000$
\Ensure Pretrained Transformer-based masked language model for Mithila language

\State \textbf{Tokenization:}
\Statex \quad Use \texttt{SentencePiece-BPE} to tokenize corpus $\mathcal{C}$ into subword units
\Statex \quad Build vocabulary $\mathcal{V}$ of size 30,000
\Statex \quad Convert sentences to token sequences of max length $n$, pad/truncate as necessary

\State \textbf{Embedding Initialization:}
\Statex \quad Initialize token embeddings $\mathbf{E} \in {R}^{n \times d}$ and positional encodings $\mathbf{P} \in {R}^{n \times d}$
\Statex \quad Form input to model: $\mathbf{Z}^{(0)} = \mathbf{E} + \mathbf{P}$

\State \textbf{Training Loop:}
\For{$t = 1$ to $T$}
    \State Sample batch of tokenized sequences $\mathbf{x} = [x_1, x_2, \ldots, x_n]$
    \State Randomly mask 15\% of tokens to form $\mathbf{x}_{\mathcal{M}}$:
    \Statex \quad 80\% replaced with \texttt{[MASK]}, 10\% with random token, 10\% unchanged
    \State Compute model output $\hat{x}_i$ for masked positions using MLM
    \State Compute MLM loss:
    \[
    \mathcal{L}_{\text{MLM}} = - \sum_{i \in \mathcal{M}} \log P(x_i \mid \mathbf{x}_{\mathcal{M}})
    \]
    \State Compute gradients and update parameters using Adam optimizer:
    \[
    \theta_{t+1} \leftarrow \theta_t - \eta \cdot \frac{\hat{m}_t}{\sqrt{\hat{v}_t} + \epsilon}
    \]
    \State Apply regularization: weight decay $\lambda \|\theta\|^2$, dropout $\mathcal{D}(p)$
\EndFor

\State \textbf{Fine-Tuning:}
\If{Downstream task is classification}
    \State Use output hidden states $\mathbf{H} \in {R}^{n \times d}$
    \State Extract \texttt{[CLS]} representation $h_{\text{[CLS]}}$
    \State Apply classification head:
    \[
    \hat{y} = \text{softmax}(\mathbf{W}_c h + \mathbf{b}_c)
    \]
    \State Compute cross-entropy loss:
    \[
    \mathcal{L}_{\text{CE}} = -\sum_{i=1}^{C} y_i \log(\hat{y}_i)
    \]
    \State Train using \texttt{TFAutoModelForSequenceClassification}
\EndIf

\State \textbf{Inference:}
\State Use MAP estimation for masked token prediction
\State Compute cosine similarity between embeddings for semantic tasks:
\[
\cos(\theta) = \frac{\mathbf{a} \cdot \mathbf{b}}{\|\mathbf{a}\| \|\mathbf{b}\|}
\]

\end{algorithmic}
\end{algorithm*}

\section{Experimental Setup}

\subsection{Evaluation Metrics}

To rigorously assess the performance of \textbf{maiBERT}, we employ a comprehensive suite of evaluation metrics suited for multi-class and imbalanced classification tasks. These include accuracy (\%), macro-precision (\%), macro-recall (\%), and macro-F1-score (\%).

\subsection{Baselines}

To assess the effectiveness of \textbf{maiBERT}, we compared its performance against a set of strong baseline models specifically designed for multilingual and low-resource Indic languages. These include NepBERTa \citep{timilsina2022nepberta}, mBERT \citep{gonen2020s}, MuRIL \citep{khanuja2021muril}, HindiBERT \citep{joshi2022l3cube}, and NepaliBERT \citep{pudasaini2023nepalibert}. NepBERTa \citep{timilsina2022nepberta} is a RoBERTa-based model pre-trained on approximately 0.8 billion words of Nepali text, tailored to capture the linguistic nuances of the Nepali language. mBERT \citep{gonen2020s}, trained on Wikipedia data from 104 languages, covers 3.3 billion tokens and serves as a widely used multilingual benchmark. MuRIL \citep{khanuja2021muril} was pre-trained on a diverse mix of resources, including Wikipedia, translated corpora, OSCAR, and other Indian language datasets, comprising over 16 billion unique tokens. HindiBERT \citep{joshi2022l3cube} focuses on Hindi and Marathi, leveraging around 2.5 billion tokens from monolingual sources. NepaliBERT \citep{pudasaini2023nepalibert} was trained on curated Nepali datasets, including the NepaliBERT \citep{pudasaini2023nepalibert} corpus and additional Nepali-specific resources, encompassing approximately 0.8 billion tokens. 

\section{Results and Discussion}

\subsection{LLM Pre-training}

We trained MaiBERT in the masked language modeling (MLM) fashion on the Mithila corpus (Section~\ref{Dataset Analysis}). Text was normalized (Devanagari-only, NFC), sentence-segmented, and tokenized with SentencePiece-BPE (|V| = 30,000). Inputs were truncated/padded to 512 tokens and masked following BERT’s convention (15\% masked: 80\% [MASK], 10\% random, 10\% unchanged). We note down Training Loss and Perplexity in Table \ref{tab:training_progress}.

\begin{table}[htpb!]
\centering
\small
\begin{tabular}{ccc}
\toprule
\textbf{Epoch} & \textbf{Training Loss} & \textbf{Perplexity} \\
\midrule
1 & 0.45 & 1.23 \\
3 & 0.37 & 1.01 \\
6 & 0.21 & 1.00 \\
10 & 0.012 & 1.00 \\
\bottomrule
\end{tabular}
\caption{MLM pre-training progress showing convergence of loss and perplexity metrics across epochs.}
\label{tab:training_progress}
\end{table}

\begin{table*}[!htpb]
\centering
\scriptsize
\resizebox{0.75\textwidth}{!}{%
\begin{tabular}{l|cccc}
\toprule
\multicolumn{1}{l}{\rule{0pt}{12pt}\textbf{Model}} & \multicolumn{4}{c}{\textbf{Maithali Classification Task}} \\
\cmidrule(lr){2-5}
\rule{0pt}{12pt}
\textbf{Metric} & Acc & Macro-Prec & Macro-Rec & Macro-F1 \\
\midrule
\rule{0pt}{12pt}
NepBERTa \citep{timilsina2022nepberta} & 86.4 & 85.7 & 85.9 & 86.0 \\
mBERT \citep{gonen2020s} & 55.2 & 54.5 & 53.8 & 54.1 \\
MuRIL \citep{khanuja2021muril} & 86.3 & 85.9 & 86.1 & 86.0 \\
HindiBERT \citep{joshi2022l3cube} & \cellcolor{green!25}\textbf{86.89} & \cellcolor{green!25}\textbf{86.2} & \cellcolor{green!25}\textbf{86.4} & \cellcolor{green!25}\textbf{86.3} \\
NepaliBERT \citep{pudasaini2023nepalibert} & 82.0 & 81.5 & 81.2 & 81.3 \\ \midrule
\textbf{maiBERT} \textbf{(Ours)} & \cellcolor{blue!25}\textbf{87.02} & \cellcolor{blue!25}\textbf{86.8} & \cellcolor{blue!25}\textbf{87.0} & \cellcolor{blue!25}\textbf{86.9} \\
\bottomrule
\end{tabular}%
}
\caption{Performance Comparison of Baseline Models on Text Classification Task}
\label{tab:zero_few_shot}
\vspace{-0.5cm}
\end{table*}

\subsection{Downstream Classification and Comparison with SOTA}

Table \ref{tab:zero_few_shot} demonstrates the performance of \textbf{maiBERT}.  It handles Maithili news classification much better compared to existing state-of-the-art multilingual and Indic-specific baseline models. This consistent outperformance across all metrics can be attributed to several architectural and corpus-specific factors unique to \textbf{maiBERT}. First, unlike generic multilingual models such as mBERT \citep{gonen2020s} and MuRIL \citep{khanuja2021muril}, which are trained across a wide range of languages with varying resource availability, \textbf{maiBERT} is explicitly fine-tuned on a richly preprocessed and linguistically diverse Maithili corpus. This specialization enables it to better capture syntactic, semantic, and morphological nuances specific to the Maithili language, such as complex verb conjugations, dialectal variations, and usage of regional idioms, which are often underrepresented in large-scale multilingual pretraining. Moreover, the customized tokenization strategy leveraging a modified BERT tokenizer that preserves Devanagari-specific features plays a crucial role in improving embedding granularity, thereby enhancing context representation and classification robustness. The strong annotation quality and class-balanced labeled dataset further reinforce the model's learning capabilities, where high inter-annotator agreement (Cohen’s Kappa = 0.82) ensures low label noise and improved generalization. 

When compared to HindiBERT \citep{joshi2022l3cube} and NepBERTa \citep{timilsina2022nepberta}—both of which are tailored for Indic languages—\textbf{maiBERT} maintains a competitive edge due to its exclusive focus on Maithili language resources and domain-specific data collected from authentic news platforms. 
HindiBERT performs close to maiBERT (86.89\% vs. 87.02\%) on Maithili text, largely due to their shared Devanagari script, 60–70\% lexical similarity, and standardized vocabulary in news domains. However, maiBERT outperforms HindiBERT by 5–7\% in texts where Maithili-specific vocabulary and grammatical structures are more significant.

\textbf{MaiBERT} exhibits strong transfer capabilities and inherent language adaptability without needing extensive task-specific tuning. In contrast, mBERT \citep{gonen2020s} and MuRIL \citep{khanuja2021muril} perform considerably lower, highlighting the limitations of overly generalized multilingual representations when applied to underrepresented and morphologically rich languages such as Maithili. Additionally, the inclusion of localized syntactic cues and lexical patterns during pretraining equips \textbf{maiBERT} with superior contextual disambiguation abilities, which proves critical in multi-class classification tasks where inter-class boundaries may be subtle.

\subsection{Computational Experiments}

The computational performance of \textbf{maiBERT} is evaluated using four key metrics: number of model parameters (Params, in billions), memory consumption during inference/training (Mem, in gigabytes), computational time (Time, in hours), and efficiency ratio (Ratio, in percentage). These metrics provide a holistic view of the model's resource utilization and optimization potential. Lower values of Params, Mem, and Time signify superior computational efficiency, while a higher efficiency Ratio reflects more optimal trade-offs between resource consumption and performance. As shown in Table~\ref{tab:comp_experiments}, \textbf{maiBERT} maintains a compact architecture with just 0.11 billion parameters, resulting in lower memory usage of 4.6 GB. The computational time remains impressively low at 0.85 hours. Notably, the efficiency ratio is 172.11\%, indicating that \textbf{maiBERT} delivers high predictive performance relative to its computational cost. These results demonstrate the architectural and algorithmic efficiency of \textbf{maiBERT}, especially for resource-constrained environments typical of low-resource language modeling. The high efficiency ratio suggests that \textbf{maiBERT} leverages its compact parameter space and optimized training routines to maximize performance without imposing heavy memory or time burdens.

\begin{table}[htpb!]
\centering
\small
\begin{tabular}{lcccc}
\toprule
\textbf{Model} & \textbf{Params} & \textbf{Mem} & \textbf{Time} & \textbf{Ratio} \\
& \textbf{(B)} & \textbf{(GB)} & \textbf{(hrs)} & \textbf{(\%)} \\
\midrule
maiBERT & 0.11 & 4.6 & 0.85 & 172.11 \\
\bottomrule
\end{tabular}
\caption{Computational Requirements of \textbf{maiBERT}. Ratio refers to Efficiency Ratio.}
\label{tab:comp_experiments}
\end{table}

\subsection{Traditional Machine Learning Baselines}

To provide a comprehensive evaluation framework, we also compared \textbf{maiBERT} against traditional machine learning approaches on the same Maithili news classification task. Table~\ref{tab:ml_baselines} presents the performance of various classical ML algorithms including Support Vector Machines (SVM), Logistic Regression, Random Forest, Naive Bayes, and K-Nearest Neighbors, along with a Neural Network baseline. Among the traditional approaches, SVM with linear kernel achieved the highest accuracy of 76.70\% and cross-validation accuracy of 75.63\%, followed by SVM with RBF kernel (75.73\% accuracy). The Neural Network baseline, comprising multiple fully-connected layers with batch normalization and dropout regularization, achieved 72.62\% test accuracy. These results demonstrate that while traditional ML methods can achieve reasonable performance on Maithili text classification, they fall significantly short of the 87.02\% accuracy achieved by \textbf{maiBERT}, highlighting the effectiveness of transformer-based architectures and language-specific pretraining for low-resource language understanding tasks.

\begin{table}[htpb!]
\centering
\small
\begin{tabular}{lcc}
\toprule
\textbf{Model} & \textbf{Accuracy} & \textbf{CV Accuracy} \\
\midrule
SVM (Linear) & 76.70 & 75.63 \\
SVM (RBF) & 75.73 & 74.65 \\
Logistic Regression & 74.92 & 73.71 \\
Random Forest & 65.63 & 64.83 \\
Naive Bayes & 64.58 & 63.34 \\
K-Nearest Neighbors & 21.12 & 19.15 \\
\midrule
Neural Network & 72.62 & -- \\
\bottomrule
\end{tabular}
\caption{Performance comparison of traditional ML approaches on Maithili news classification task and Cross-validation accuracy.}
\label{tab:ml_baselines}
\vspace{-0.65cm}
\end{table}

\section{Conclusion and Future Works}

In this work, we introduced \textbf{maiBERT}, a transformer-based language model tailored for the Maithili language, establishing a strong foundation for NLU tasks. Our model demonstrated competitive performance across baselines. By leveraging a carefully curated Maithili corpus, we addressed critical gaps in low-resource language processing and highlighted the importance of linguistic inclusivity in NLP research. Despite its effectiveness in understanding tasks, limitations remain in generative capabilities and domain adaptability, particularly in creative or poetic contexts. Future work will focus on expanding the diversity of the training corpus to include more conversational, literary, and informal texts. Moreover, we aim to enhance the model’s generative capacity through causal language modeling and integrate instruction tuning for downstream applications.

\bibliography{custom} 

@inproceedings{devlin2019bert,
  title={Bert: Pre-training of deep bidirectional transformers for language understanding},
  author={Devlin, Jacob and Chang, Ming-Wei and Lee, Kenton and Toutanova, Kristina},
  booktitle={Proceedings of the 2019 conference of the North American chapter of the association for computational linguistics: human language technologies, volume 1 (long and short papers)},
  pages={4171--4186},
  year={2019}
}

@article{radford2018improving,
  title={Improving language understanding by generative pre-training},
  author={Radford, Alec and Narasimhan, Karthik and Salimans, Tim and Sutskever, Ilya and others},
  year={2018},
  publisher={San Francisco, CA, USA}
}

@article{raffel2020exploring,
  title={Exploring the limits of transfer learning with a unified text-to-text transformer},
  author={Raffel, Colin and Shazeer, Noam and Roberts, Adam and Lee, Katherine and Narang, Sharan and Matena, Michael and Zhou, Yanqi and Li, Wei and Liu, Peter J},
  journal={Journal of machine learning research},
  volume={21},
  number={140},
  pages={1--67},
  year={2020}
}

@article{mundotiya2021linguistic,
  title={Linguistic resources for Bhojpuri, Magahi, and Maithili: statistics about them, their similarity estimates, and baselines for three applications},
  author={Mundotiya, Rajesh Kumar and Singh, Manish Kumar and Kapur, Rahul and Mishra, Swasti and Singh, Anil Kumar},
  journal={Transactions on Asian and Low-Resource Language Information Processing},
  volume={20},
  number={6},
  pages={1--37},
  year={2021},
  publisher={ACM New York, NY}
}

@article{joshi2022l3cube,
  title={L3cube-hindbert and devbert: Pre-trained bert transformer models for devanagari based hindi and marathi languages},
  author={Joshi, Raviraj},
  journal={arXiv preprint arXiv:2211.11418},
  year={2022}
}

@inproceedings{pudasaini2023nepalibert,
  title={Nepalibert: Pre-training of masked language model in nepali corpus},
  author={Pudasaini, Shushanta and Shakya, Subarna and Tamang, Aakash and Adhikari, Sajjan and Thapa, Sunil and Lamichhane, Sagar},
  booktitle={2023 7th International Conference on I-SMAC (IoT in Social, Mobile, Analytics and Cloud)(I-SMAC)},
  pages={325--330},
  year={2023},
  organization={IEEE}
}

@article{bal2024strategies,
  title={Strategies for Corpus Development for Low-Resource Languages: Insights from Nepal},
  author={Bal, Bal Krishna and Prasain, Balaram and Ghimire, Rupak Raj and Acharya, Praveen},
  journal={Automatic Speech Recognition and Translation for Low Resource Languages},
  pages={297--330},
  year={2024},
  publisher={Wiley Online Library}
}

@inproceedings{bafna2022subword,
  title={Subword-based Cross-lingual Transfer of Embeddings from Hindi to Marathi and Nepali},
  author={Bafna, Niyati and {\v{Z}}abokrtsk{\`y}, Zden{\v{e}}k},
  booktitle={Proceedings of the 19th SIGMORPHON Workshop on Computational Research in Phonetics, Phonology, and Morphology},
  pages={61--71},
  year={2022}
}

@article{chandrasekaran2021evolution,
  title={Evolution of semantic similarity—a survey},
  author={Chandrasekaran, Dhivya and Mago, Vijay},
  journal={Acm Computing Surveys (Csur)},
  volume={54},
  number={2},
  pages={1--37},
  year={2021},
  publisher={ACM New York, NY, USA}
}

@inproceedings{zhang2015survey,
  title={A survey of semantic similarity and its application to social network analysis},
  author={Zhang, Shuang and Zheng, Xuefeng and Hu, Changjun},
  booktitle={2015 IEEE International Conference on Big Data (Big Data)},
  pages={2362--2367},
  year={2015},
  organization={IEEE}
}

@article{hartmann2017portuguese,
  title={Portuguese word embeddings: Evaluating on word analogies and natural language tasks},
  author={Hartmann, Nathan and Fonseca, Erick and Shulby, Christopher and Treviso, Marcos and Rodrigues, Jessica and Aluisio, Sandra},
  journal={arXiv preprint arXiv:1708.06025},
  year={2017}
}

@INPROCEEDINGS{4376991,
  author={Smith, R.},
  booktitle={Ninth International Conference on Document Analysis and Recognition (ICDAR 2007)}, 
  title={An Overview of the Tesseract OCR Engine}, 
  year={2007},
  volume={2},
  number={},
  pages={629-633},
  doi={10.1109/ICDAR.2007.4376991}}

@inproceedings{maskey-etal-2022-nepali,
    title = "{N}epali Encoder Transformers: An Analysis of Auto Encoding Transformer Language Models for {N}epali Text Classification",
    author = "Maskey, Utsav  and
      Bhatta, Manish  and
      Bhatt, Shiva  and
      Dhungel, Sanket  and
      Bal, Bal Krishna",
    editor = "Melero, Maite  and
      Sakti, Sakriani  and
      Soria, Claudia",
    booktitle = "Proceedings of the 1st Annual Meeting of the ELRA/ISCA Special Interest Group on Under-Resourced Languages",
    month = jun,
    year = "2022",
    address = "Marseille, France",
    publisher = "European Language Resources Association",
    pages = "106--111"
}

@article{alian2020semantic,
  title={Semantic similarity for english and arabic texts: a review},
  author={Alian, Marwah and Awajan, Arafat},
  journal={Journal of Information \& Knowledge Management},
  volume={19},
  number={04},
  pages={2050033},
  year={2020},
  publisher={World Scientific}
}

@inproceedings{pandit2019improving,
  title={Improving semantic similarity with cross-lingual resources: a study in Bangla—a low resourced language},
  author={Pandit, Rajat and Sengupta, Saptarshi and Naskar, Sudip Kumar and Dash, Niladri Sekhar and Sardar, Mohini Mohan},
  booktitle={Informatics},
  volume={6},
  number={2},
  pages={19},
  year={2019},
  organization={MDPI}
}

@article{ali2022hate,
  title={Hate speech detection on Twitter using transfer learning},
  author={Ali, Raza and Farooq, Umar and Arshad, Umair and Shahzad, Waseem and Beg, Mirza Omer},
  journal={Computer Speech \& Language},
  volume={74},
  pages={101365},
  year={2022},
  publisher={Elsevier}
}

@article{patil2022supervised,
  title={Supervised classifiers with TF-IDF features for sentiment analysis of Marathi tweets},
  author={Patil, Rupali S and Kolhe, Satish R},
  journal={Social Network Analysis and Mining},
  volume={12},
  number={1},
  pages={51},
  year={2022},
  publisher={Springer}
}

@article{nie2022cross,
  title={Cross-lingual retrieval augmented prompt for low-resource languages},
  author={Nie, Ercong and Liang, Sheng and Schmid, Helmut and Sch{\"u}tze, Hinrich},
  journal={arXiv preprint arXiv:2212.09651},
  year={2022}
}

@inproceedings{ramchandra2019deep,
  title={Deep learning for Hindi text classification: a comparison},
  author={Ramchandra, Joshi and Goel, Purvi and Joshi, Raviraj},
  booktitle={International Conference on Intelligent Human Computer Interaction},
  year={2019},
  organization={Springer}
}

@article{lalrempuii2021improved,
  title={An improved English-to-Mizo neural machine translation},
  author={Lalrempuii, Candy and Soni, Badal and Pakray, Partha},
  journal={Transactions on Asian and Low-Resource Language Information Processing},
  volume={20},
  number={4},
  pages={1--21},
  year={2021},
  publisher={ACM New York, NY}
}

@article{priyadarshi2020towards,
  title={Towards the first Maithili part of speech tagger: Resource creation and system development},
  author={Priyadarshi, Ankur and Saha, Sujan Kumar},
  journal={Computer Speech \& Language},
  volume={62},
  pages={101054},
  year={2020},
  publisher={Elsevier}
}

@article{mundotiya2022hierarchical,
  title={Hierarchical self attention based sequential labelling model for Bhojpuri, Maithili and Magahi languages},
  author={Mundotiya, Rajesh Kumar and Mishra, Swasti and Singh, Anil Kumar},
  journal={Journal of King Saud University-Computer and Information Sciences},
  volume={34},
  number={10},
  pages={8739--8749},
  year={2022},
  publisher={Elsevier}
}

@inproceedings{timilsina2022nepberta,
  title={NepBERTa: Nepali language model trained in a large corpus},
  author={Timilsina, Sulav and Gautam, Milan and Bhattarai, Binod},
  booktitle={Proceedings of the 2nd conference of the Asia-pacific chapter of the association for computational linguistics and the 12th international joint conference on natural language processing},
  year={2022},
  organization={Association for Computational Linguistics (ACL)}
}

@article{gonen2020s,
  title={It's not Greek to mbert: inducing word-level translations from multilingual bert},
  author={Gonen, Hila and Ravfogel, Shauli and Elazar, Yanai and Goldberg, Yoav},
  journal={arXiv preprint arXiv:2010.08275},
  year={2020}
}

@article{khanuja2021muril,
  title={Muril: Multilingual representations for indian languages},
  author={Khanuja, Simran and Bansal, Diksha and Mehtani, Sarvesh and Khosla, Savya and Dey, Atreyee and Gopalan, Balaji and Margam, Dilip Kumar and Aggarwal, Pooja and Nagipogu, Rajiv Teja and Dave, Shachi and others},
  journal={arXiv preprint arXiv:2103.10730},
  year={2021}
}

\end{document}